\documentclass[runningheads]{llncs}
\usepackage{graphicx}
\usepackage{hyperref}
\usepackage{subfig}
\usepackage{amsmath}
\usepackage{amsfonts}
\usepackage{algorithmic}
\usepackage{enumitem}
\begin{document}
\title{Probabilistic Motion Modeling from Medical Image Sequences: Application to Cardiac Cine-MRI}
%\title{ \thanks{Supported by organization x.}}

\author{Julian Krebs\inst{1,2} \and Tommaso Mansi\inst{1} \and Nicholas Ayache\inst{2} \and Herv\'{e} Delingette\inst{2}}

%Second Author\inst{2,3}\orcidID{1111-2222-3333-4444} \and

\authorrunning{J. Krebs et al.}

\institute{Siemens Healthineers, Digital Services, Digital Technology and Innovation, Princeton, NJ, 08540 USA \and Universit\'{e} C\^{o}te d'Azur, Inria, Epione Team, Sophia Antipolis, 06902 France}

\maketitle              % typeset the header of the contribution
%
%\vspace{-8pt}
\begin{abstract}
We propose to learn a probabilistic motion model from a sequence of images. Besides spatio-temporal registration, our method offers to predict motion from a limited number of frames, useful for temporal super-resolution. The model is based on a probabilistic latent space and a novel temporal dropout training scheme. This enables simulation and interpolation of realistic motion patterns given only one or any subset of frames of a sequence. The encoded motion also allows to be transported from one subject to another without the need of inter-subject registration. 
An unsupervised generative deformation model is applied within a temporal convolutional network which leads to a diffeomorphic motion model -- encoded as a low-dimensional motion matrix. Applied to cardiac cine-MRI sequences, we show improved registration accuracy and spatio-temporally smoother deformations compared to three state-of-the-art registration algorithms. Besides, we demonstrate the model's applicability to motion transport by simulating a pathology in a healthy case. Furthermore, we show an improved motion reconstruction from incomplete sequences compared to linear and cubic interpolation.

%\keywords{Probabilistic Motion Model \and Tracking \and Temporal Super-Resolution \and Diffeomorphic Registration \and Variational Autoencoder.}
%\vspace{-5pt}
\end{abstract}

\section{Introduction}
%\vspace{-5pt}
In medical imaging, an important task is to analyze temporal image sequences to understand physiological processes of the human body. Dynamic organs, such as the heart or lungs, are of particular interest to study as detected motion patterns are helpful for the diagnosis and treatment of diseases. Moreover, recovering the motion pattern allows to track anatomical structures, to compensate for motion, to do temporal super-resolution and motion simulation.

Motion is typically studied by computing pairwise deformations -- the registration of each of the images in a sequence with a target image. The resulting dense deformation fields track moving structures from the beginning to the end of the sequence. Providing an invertible and smooth transformation, diffeomorphic registration algorithms such as the SyN algorithm \cite{avants2008symmetric}, the LCC-demons \cite{lorenzi2013lcc} or recent learning-based algorithms \cite{dalca2018unsupervised,krebs2019} are especially suited for the registration of sequential images. One difficulty is to acquire temporally smooth deformations that are fundamental for tracking. That is why registration algorithms with a temporal regularizer have been proposed \cite{de2012temporal,metz2011nonrigid,qin2018joint,shi2013temporal}. %These methods are based on temporal B-Spline free form deformations \cite{de2012temporal} or . 
In the computer vision community, temporal video super-resolution and motion compensation are of related interest \cite{caballero2017real}.

However, while these methods produce accurate dense deformations, they do not aim to extract intrinsic motion parameters crucial for building a comprehensive motion model useful for analysis tasks such as motion classification or simulation. Roh\'{e} et al.~\cite{rohe2018low} proposed  a  parameterization, the Barycentric Subspaces, as a regularizer for cardiac motion tracking. Yang et al.~\cite{yang2011prediction} generated a motion prior using manifold learning from low-dimensional shapes.

We propose to learn a probabilistic motion model from image sequences directly. Instead of defining a parameterization explicitly or learning from pre-processed shapes, our model captures relevant motion features in a low-dimensional motion matrix in a generic but data-driven way. This learned latent space can be used to fill gaps of missing frames (motion reconstruction), to predict the next frames in the sequence or to generate an infinite number of new motion patterns given only one image (motion simulation). Motion can be also transported by applying the motion matrix on an image of another subject. 
%This feature offers novel potential applications such as quicker data acquisition through sequence reconstruction from a limited number of frames, sequential motion estimation from only two given frames or data augmentation.

The probabilistic motion encoding is learned by generalizing a pair-wise registration method \cite{krebs2019} based on Bayesian inference \cite{kingma2014semi} using a temporal regularizer with explicit time dependence. Furthermore, to enforce temporal consistency, we introduce a novel self-supervised training scheme called temporal dropout sampling. The framework is learned in an unsupervised fashion from image sequences of varying lengths. Smooth, diffeomorphic and symmetric deformations are ensured by applying an exponentiation layer, spatio-temporal regularization and a symmetric local cross-correlation metric. Besides motion simulation, the model demonstrates state-of-the-art registration results for diffeomorphic tracking of cardiac cine-MRI. The main contributions are as follows:
%\vspace{-5pt}
\begin{itemize}[noitemsep]
\item An unsupervised probabilistic motion model learned from image sequences
\item A generative model using explicit time-dependent temporal convolutional networks trained with self-supervised temporal dropout sampling
\item Demonstration of cardiac motion tracking, simulation, transport and temporal super-resolution
\end{itemize}
%\vspace{-10pt}

\section{Methods}
\begin{figure}[tb]
\centering
\begin{minipage}{.51\linewidth}
\subfloat[]{\includegraphics[trim=0 203 690 0,clip,width=1\linewidth]{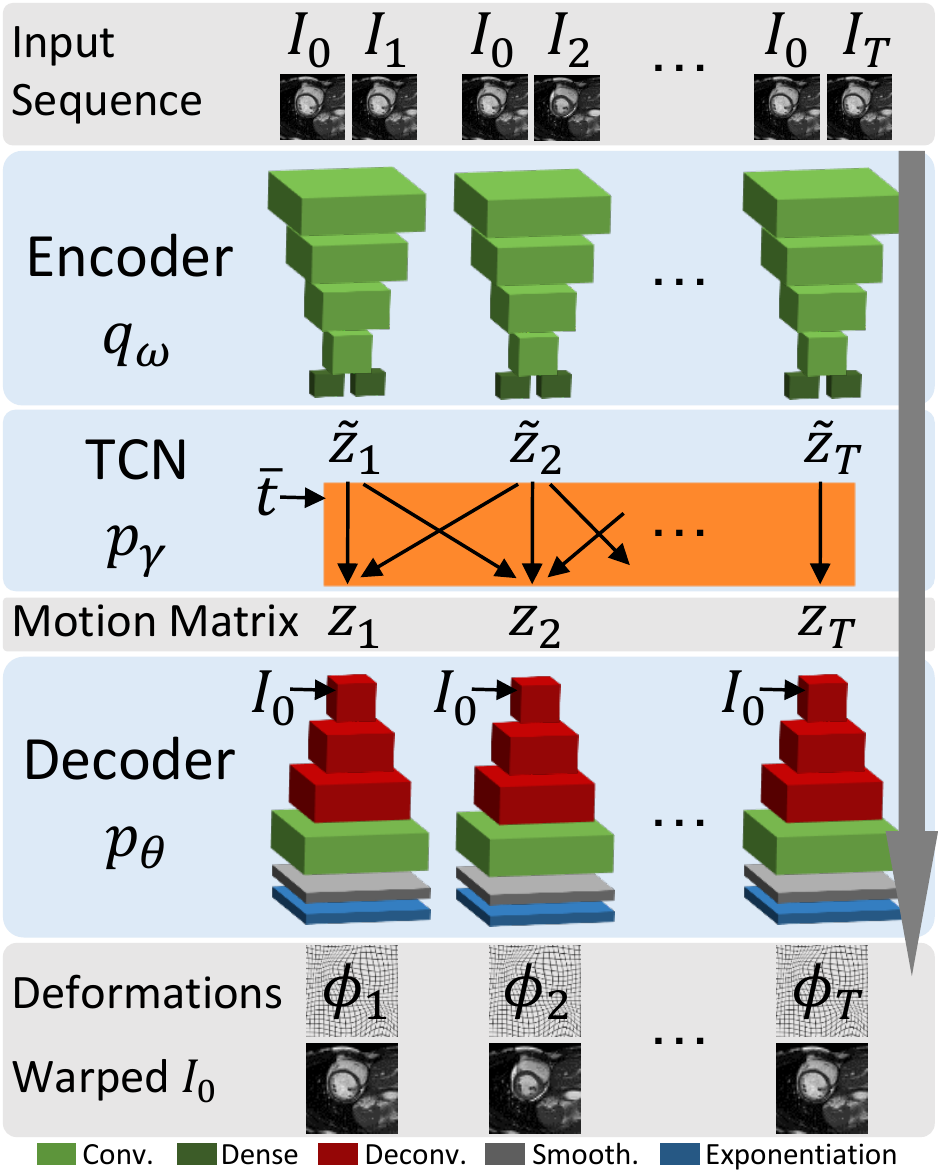}\label{model_fig}}
\end{minipage}
\begin{minipage}{.43\linewidth}
\subfloat[]{\includegraphics[trim=0 330 640 0,clip,width=1\linewidth]{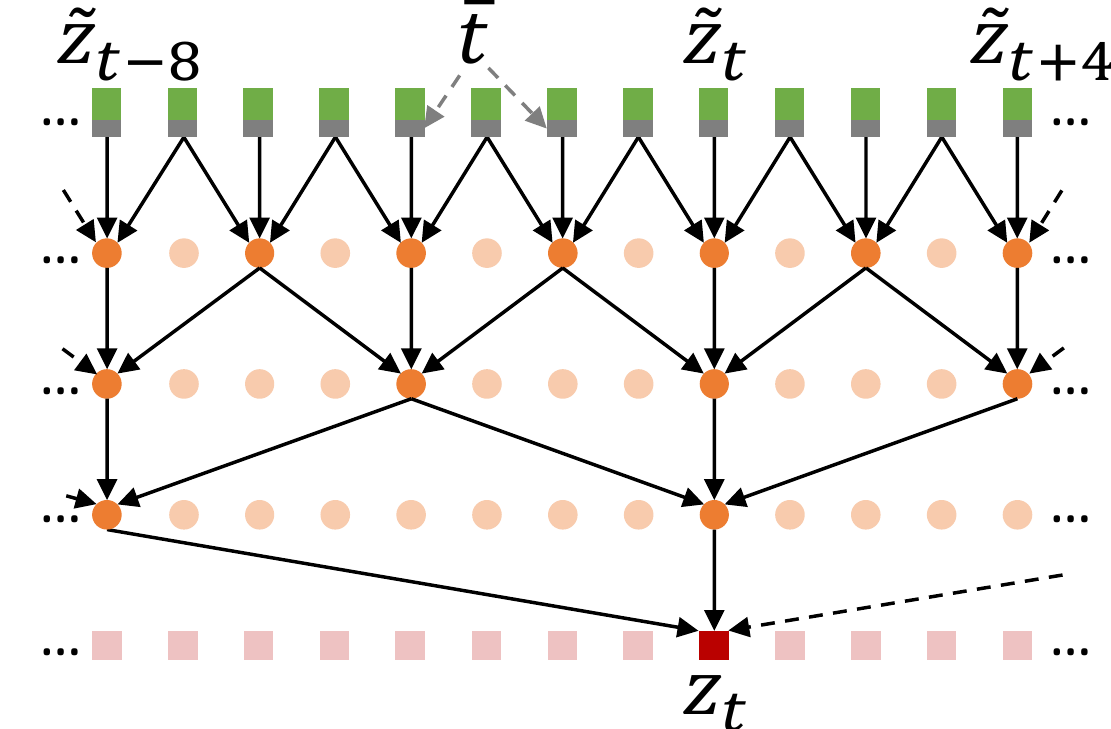}\label{tcn_fig}}\\
\subfloat[]{\includegraphics[trim=0 388 715 0,clip,width=1\linewidth]{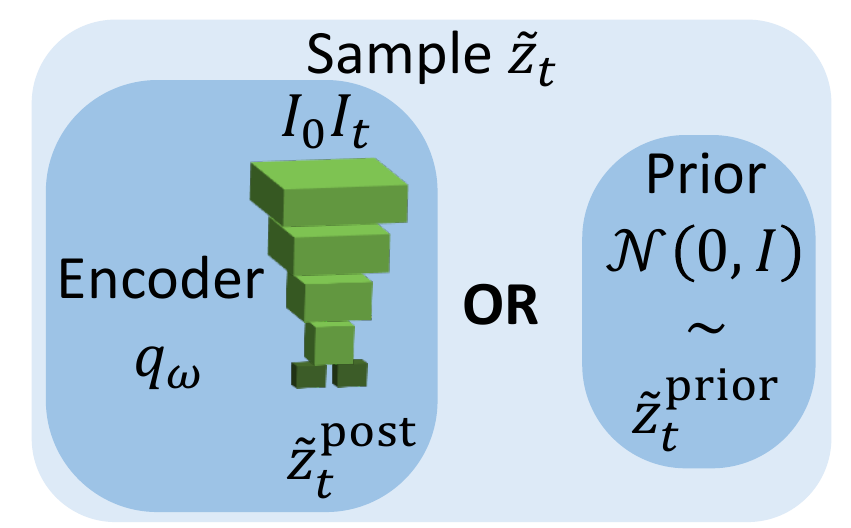}\label{sampling_fig}}
\end{minipage}
%\vspace{-8pt}
\caption{\small{Probabilistic motion model (a): The encoder $q_\omega$ projects the image pair $(I_0,I_t)$ to a low-dimensional deformation encoding $\tilde{z}_t$ from which the temporal convolutional network $p_\gamma$ (b) constructs the motion matrix $z\in \mathbb{R}^{d\times T}$ conditioned on the normalized time $\bar{t}$. The decoder $p_\theta$ maps the motion matrix to the deformations $\phi_t$. The temporal dropout sampling procedure (c) randomly chooses to sample $\tilde{z}_t$ either from the encoder $q_\omega$ or the prior distribution.}}
%\vspace{-10pt}
\end{figure}

%\vspace{-5pt}
The motion observed in an image sequence with $T+1$ frames is typically described by deformation fields $\phi_t$ between a moving image $I_0$ and the fixed images $I_t$ with $t\in[1,T]$. Inspired by the probabilistic deformation model of \cite{krebs2019} based on conditional variational autoencoder (CVAE) \cite{kingma2014semi}, we define a motion model for temporal sequences. The model is conditioned on the moving image and parameterizes the set of  diffeomorphisms $\phi_t$ in a low-dimensional probabilistic space, the motion matrix  $z\in \mathbb{R}^{d\times T}$, where $d$ is the size of the deformation encoding per image pair. Each column's $z_t$-code corresponds to the deformation $\phi_t$. To take temporal dependencies into account, $z_t$ is conditioned on all past and future time steps. To learn this temporal regularization directly from data, we apply Temporal Convolutional Networks \cite{koltun2018} with explicit time dependence and temporal dropout sampling enforcing the network to fill time steps by looking at given past and future deformations. An illustration of the model is shown in Fig.~\ref{model_fig}.

%\vspace{-5pt}
\subsubsection{Probabilistic Motion Model}
Our motion model consists of three distributions. First, the encoder $q_\omega(\tilde{z}_t|I_0,I_t)$ maps each of the image pairs $(I_0, I_t)$ independently to a latent space denoted by $\tilde{z}_t \in \mathbb{R}^{d}$. Second, as the key component of temporal modeling, these latent vectors $\tilde{z}_t$ are jointly mapped to the motion matrix $z$ by conditioning them in all past and future time steps and on the normalized time $\bar{t}$: $p_\gamma(z|\tilde{z}_{1:T}, \bar{t}_{1:T})$. Finally, the decoder $p_\theta(I_t|z_t, I_0)$ aims to reconstruct the fixed image $I_t$ by warping the moving image $I_0$ with the deformation $\phi_t$. This deformation $\phi_t$ is extracted from the temporally regularized $z_t$-codes. The decoder is conditioned on the moving image by concatenating the features at each scale with down-sampled versions of $I_0$.

The distributions $q_\omega$, $p_\gamma$, $p_\theta$ are approximated by three neural networks with trainable parameters $\omega$, $\gamma$, $\theta$. During training, a lower bound on the data likelihood is maximized with respect to a prior distribution $p(\tilde{z}_t)$ of the latent space $\tilde{z}_t$ (cf.~CVAE \cite{kingma2014semi}). The prior $p(\tilde{z}_t)$ is assumed to follow a multivariate unit Gaussian distribution with spherical covariance $I$: $p(\tilde{z}_t)\sim \mathcal{N}(0,I)$. The objective function results in optimizing the expected log-likelihood $p_\theta$ and the Kullback-Leibler (KL) divergence enforcing the posterior distribution $q_\omega$ to be close to the prior $p(\tilde{z}_t)$ for all time steps:%\vspace{-5pt}
\begin{equation}\label{objective}
\sum_{t=1}^T \mathbb{E}_{z_t\sim p_\gamma(\cdot|\tilde{z}_{1:T}, \bar{t}_{1:T})} \Big[\text{log } p_\theta(I_t|z_t,I_0) \Big] - \text{KL}\left[q_\omega(\tilde{z}_t|I_0, I_t)\| p(\tilde{z}) \right].
\end{equation}
Unlike the traditional CVAE model, the temporal regularized $z_t$-code is used in the log-likelihood term $p_\theta$ instead of the $\tilde{z}_t$. We model $p_\theta$ as a symmetric local cross-correlation Boltzmann distribution with the weighting factor $\lambda$. Encoder and decoder weights are independent of the time $t$. Their network architecture consists of convolutional and deconvolutional layers with fully-connected layers for mean and variance predictions in the encoder part \cite{kingma2014semi}. We use an exponentiation layer for the stationary velocity field parameterization of diffeomorphisms \cite{krebs2019}, a linear warping layer and diffusion-like regularization with smoothing parameters $\sigma_G$ in spatial and $\sigma_T$ in temporal dimension.

%\vspace{-8pt}
\subsubsection{Temporal Convolutional Networks with Explicit Time Dependence}
Since the parameters of encoder $q_\omega$ and decoder $p_\theta$ are independent of time, the temporal conditioning $p_\gamma$ plays an important role in merging information across different time steps. In our work, this regularization is learned by Temporal Convolutional Networks (TCN). Consisting of multiple 1-D convolutional layers with increasing dilation, TCN can handle input sequences of different lengths. TCN have several advantages compared to recurrent neural networks such as a flexible receptive field and more stable gradient computations \cite{koltun2018}. 

The input of the TCN is the sequence of $\tilde{z}$ concatenated with the normalized time $\bar{t}=t/T$. Providing the normalized time explicitly, provides the network with information on where each $\tilde{z}$ is located in the sequence. This supports the learning of a motion model from data representing the same type of motion with varying sequence lengths. The output of the TCN is the regularized motion matrix $z$. We use non-causal instead of causal convolutional layers to also take future time steps into account. We follow the standard implementation using zero-padding and skip connections. Each layer contains $d$ filters. A schematic representation of our TCN is shown in Fig.~\ref{tcn_fig}. For cyclic sequences, one could use a cyclic padding instead of zero-padding, for example by linking $\tilde{z}_T$ to $\tilde{z}_0$. However, in case of cardiac cine-MRI, one can not assume the end of a sequence coincides with the beginning as 5-10\% of the cardiac cycle are often omitted \cite{bernard2018deep}. 

%\vspace{-8pt}
\subsubsection{Training with Temporal Dropout Sampling}
Using Eq.~\ref{objective} for training could lead to learning the identity transform $z\approx\tilde{z}$ in the TCN $p_\gamma$ such that deformations of the current time step are independent of past and future time steps. To avoid this and enforce the model to search for temporal dependencies during the training, we introduce the concept of temporal dropout sampling (TDS). In TDS, some of the $\tilde{z}_t$ are sampled from the prior distribution $p(\tilde{z})$ instead of only sampling from the posterior distribution $q_\omega(\tilde{z}_t|I_0, I_t)$ as typical for CVAE. At the time steps the prior has been used for sampling, the model has no knowledge of the target image $I_t$ and is forced to use the temporal connections within the TCN in order to minimize the objective.

More precisely, at each time step $t$, a sample from the prior distribution $\tilde{z}_t^{\text{prior}}\sim p(\tilde{z}_t)$ is selected instead of a posterior sample $\tilde{z}_t^{\text{post}}\sim q_\omega(\tilde{z}_t|I_0, I_t)$ using a binary Bernoulli random variable $r_t$. All independent Bernoulli random variables $r\in \mathbb{R}^T$ have the success probability $\delta$. The latent vector $\tilde{z}_t$ can be defined as:
\begin{equation}
\tilde{z}_t = r_t * \tilde{z}_t^{\text{prior}} + (1-r_t) * \tilde{z}_t^{\text{post}}.
\end{equation}
%\vspace{-2pt}
Fig.~\ref{sampling_fig} illustrates the TDS procedure. At test time, for each time step independently, one can either draw $\tilde{z}_t$ from the prior or take the encoder's prediction.

\section{Experiments}
%\vspace{-5pt}
We evaluate our motion model on 2-D cardiac MRI-cine data. First, we demonstrate accurate temporal registration by evaluating motion tracking and compensation of the cardiac sequence, taking the end-diastolic (ED) frame as the moving image $I_0$. Stabilization, is accomplished by warping all frames $I_t$ to the ED frame. Pair-wise registration results are presented for ED-ES (end-systolic) frame pairs. Second, we present motion transport, motion sampling and reconstruction with a limited number of frames.

%\vspace{-9pt}
\subsubsection{Data} We used 334 short-axis sequences acquired from different hospitals including 150 sequences from the Automatic Cardiac Diagnosis Challenge 2017 (ACDC \cite{bernard2018deep}). The remaining cases were obtained from the EU FP7-funded project MD-Paedigree (Grant Agreement 600932), mixing congenital heart diseases with healthy and pathological images from adults. The cine images were acquired in breath hold using 1R-R or 2R-R intervals with a retrospective or prospective gating. The sequence length $T$ varied from 13 to 35 frames. We used the 100 cases from ACDC that contain ED-ES segmentation information for testing while the remaining sequences were used for training. All slices were resampled with a spacing of 1.5 $\times$ 1.5 mm and cropped to a size of 128 $\times$ 128 pixels. 

%\vspace{-9pt}
\subsubsection{Implementation Details}
The encoder $q_\omega$ consisted of 4 convolutional layers with strides (2, 2, 2, 1) and dense layers of size $d$ for mean and variance estimation of the VAE. The TCN consisted of four 1-D convolutional layers with dilations (1, 2, 4, 8), \textit{same} padding, a kernel size of 3 and skip connections (cf.~Fig.~\ref{tcn_fig}). The decoder $p_\theta$ had 3 deconvolutional and 1 convolutional layer before the exponentiation and warping layers (Fig.~\ref{model_fig}).  The regularization parameters $\sigma_G$ and $\sigma_T$ were set to 3 mm respectively 1.5. The loss weighting factor $\lambda$ was chosen empirically as $6 \cdot10^4$. The deformation encoding size $d$ was set to 32. The dropout sampling probability $\delta$ was 0.5. We applied a first-order gradient-based method for stochastic optimization (Adam \cite{kingma2014adam}) with a learning rate of 0.00015 and a batch size of one. We performed data augmentation on-the-fly by randomly shifting, rotating, scaling and mirroring images. We implemented the model in Tensorflow \cite{abadi2016tensorflow} with Keras\footnote{https://github.com/fchollet/keras}. The training time was 15h on a NVIDIA GTX TITAN X GPU.

%\vspace{-3pt}
\subsection{Registration: Tracking and Motion Compensation}
%\vspace{-4pt}
\begin{figure}[tb]
\centering
\includegraphics[trim=0 114 369 0,clip,width=1\linewidth]{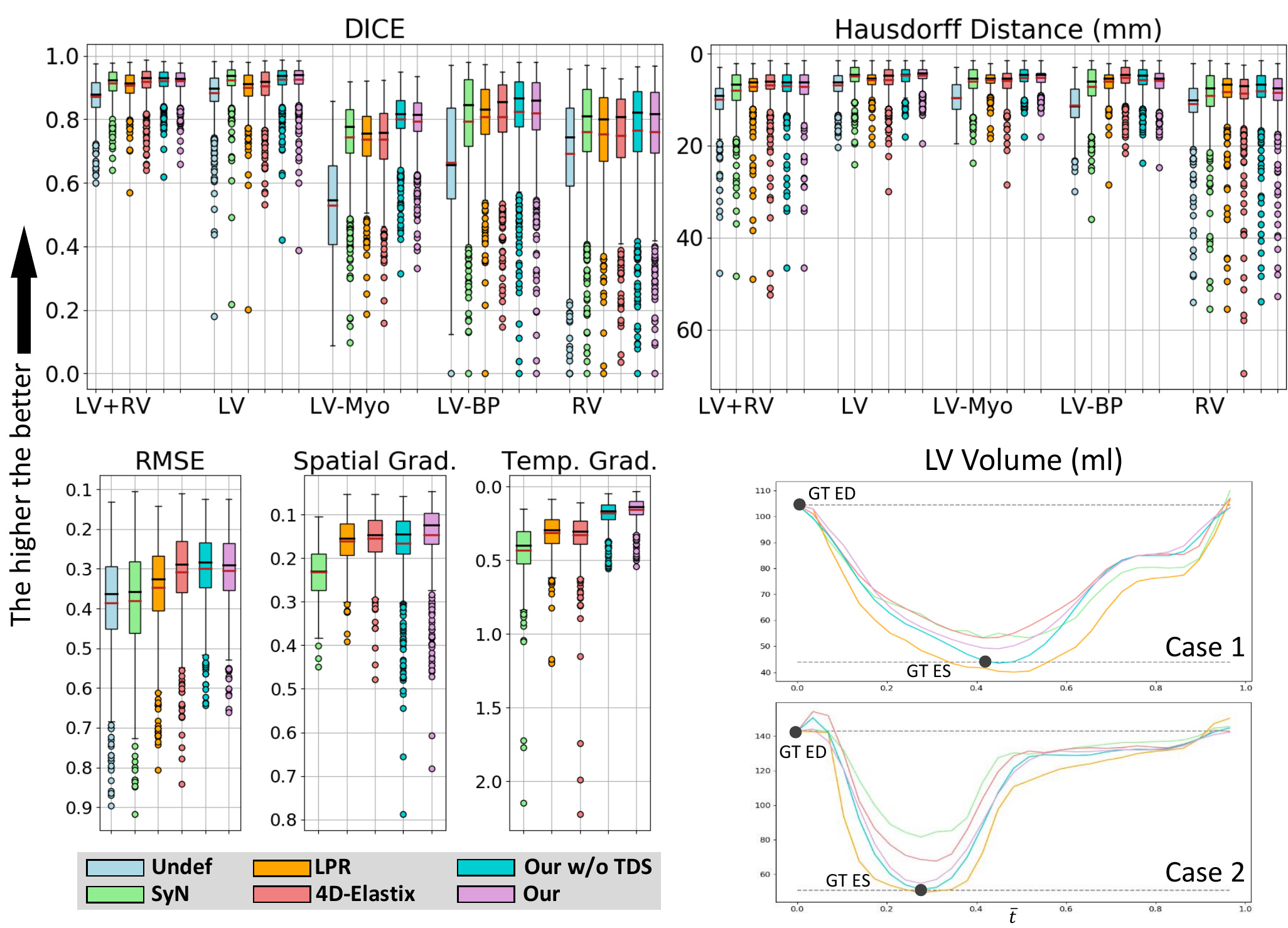}
%\vspace{-15pt}
\caption{\small{Tracking results showing RMSE, spatial and temporal gradient of the displacement fields, DICE scores and  Hausdorff distances. LV volumes in ml are shown for two test sequences (ground truth ED/ES volumes marked with points). The proposed algorithms (Our and Our w/o  TDS) show slightly higher registration accuracy and temporally smoother deformations than the state-of-the-art algorithms: SyN \cite{avants2008symmetric}, LPR \cite{krebs2019} and 4D-Elastix \cite{metz2011nonrigid}.}}\label{reg_results}
\end{figure}

\begin{figure}[tb]
\centering
%\vspace{-8pt}
\includegraphics[trim=2 210 549 0,clip,width=1\linewidth]{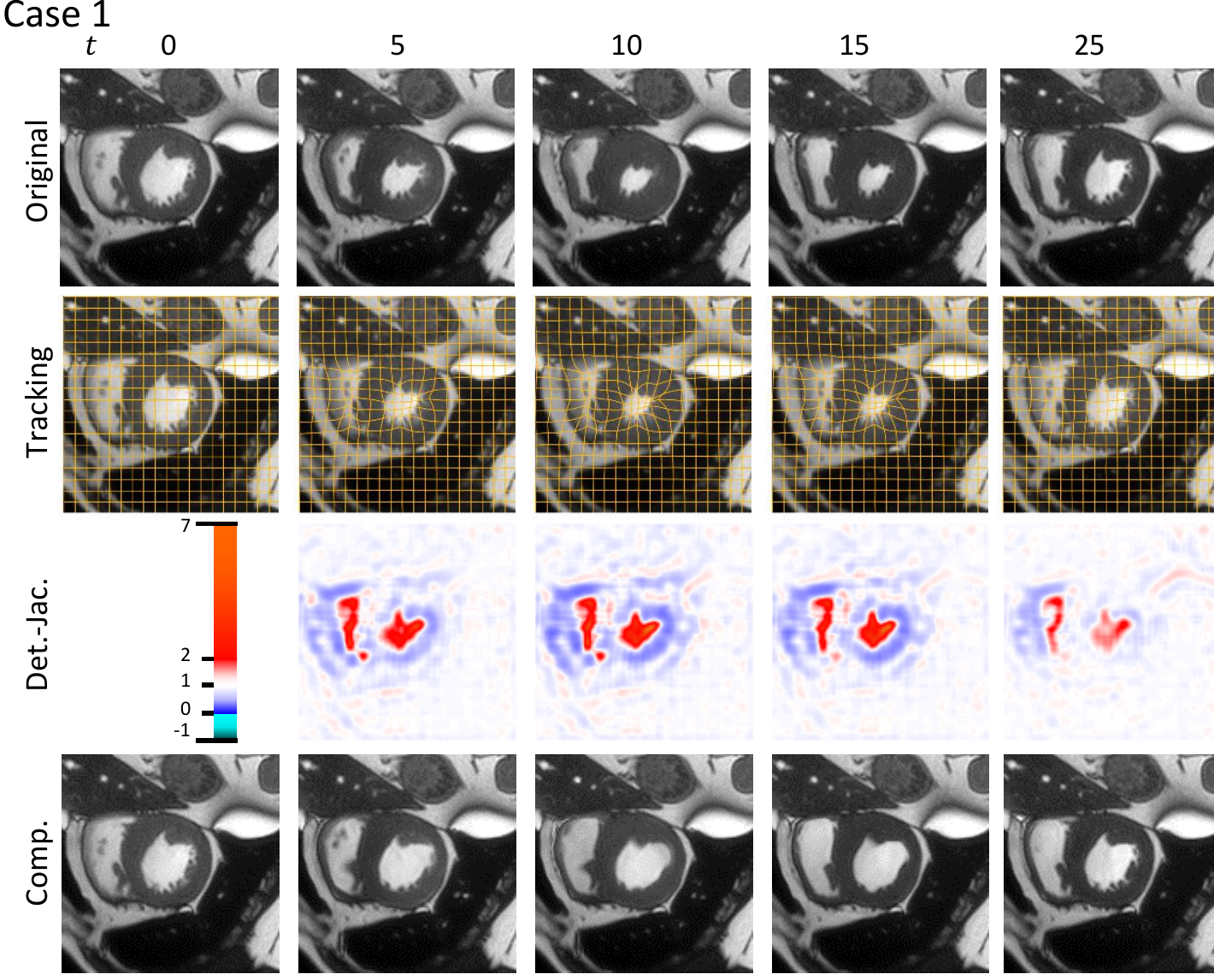}
%\vspace{-15pt}
\caption{\small{Visual results of one test sequence, showing tracking, the Jacobian determinant (Det.-Jac.) and motion compensation (Comp.).}}\label{reg_images}
%\vspace{-14pt}
\end{figure}
We compare our model with and without the temporal dropout sampling (Our w/o TDS) with three state-of-the-art methods: SyN \cite{avants2008symmetric}, the learning-based probabilistic registration (LPR, 2-D single-scale version \cite{krebs2019}) and the b-spline-based 4D algorithm in elastix \cite{metz2011nonrigid}. In contrast to the results in \cite{krebs2019}, in this work, LPR was trained in 2-D taking all images from a sequence into account, not only ED/ES pairs. The following results are reported for full sequences, except the metrics based on segmentations which are only reported on frame pairs with provided ground truth information (ED/ES pairs). 

In Fig.~\ref{reg_results}, tracking results are visualized for the test data taking sequences of all slices where segmentation in ED and ES frames are available, resulting in 677 sequences. We report the root mean square error (RMSE), the spatial (Spatial Grad.) and temporal gradient (Temp. Grad.) of the displacement fields for evaluating the smoothness of the resulting deformations. Our model shows spatially and temporally smoother deformations. %The compensation losses are defined as the sum-of-squared pixel differences of the myocardium area between ED frame and all $I_t$ (Comp. Reg.) and between all $(I_t,I_{t-1})$ pairs (Comp. Temp.) representing the temporal consistency. 
We also report DICE scores and 95\%-tile Hausdorff distances (HD in mm) on five anatomical structures: myocardium (LV-Myo) and epicardium (LV) of the left ventricle, left bloodpool (LV-BP), right ventricle (RV) and LV+RV. Note, that DICE scores and HD were evaluated on ED-ES frame pairs only. The proposed method showed improved mean DICE scores and smaller mean HD of 84.6\%, 6.2mm (w/o TDS: 84.7\%, 6.1mm) compared to SyN, LPR, 4D-Elastix with (82.7\%, 7.0mm), (82.1\%, 6.6mm) respectively (83.7\%, 6.3mm\%). %Our algorithm reached a  of 6.2mm (6.1mm w/o TDS) while SyN, LPR,  reached 7.0mm and LPR 6.6mm. 
Compared to training without temporal dropout sampling, HD and DICE scores show minimal differences, indicating that using TDS does not degrade registration accuracy while improving deformation regularity. Furthermore, only TDS offers consistent motion simulation and temporal interpolation. 

In the bottom right of Fig.~\ref{reg_results}, LV volume curves computed by warping the ED mask are plotted. One can see that the SyN algorithm underestimates big deformations. Both, the volume and the gradient metrics show smoother deformations for both versions of our motion model compared to the SyN, LPR and 4D-Elastix algorithms. Visual results for one case including tracking, determinant of Jacobians and compensated motion are shown in Fig.~\ref{reg_images}. Motion compensation was done by warping the $I_t$'s frame with inverted diffeomorphisms.

\begin{figure}[tb]
\centering
\includegraphics[trim=16 164 218 0,clip,width=1\linewidth]{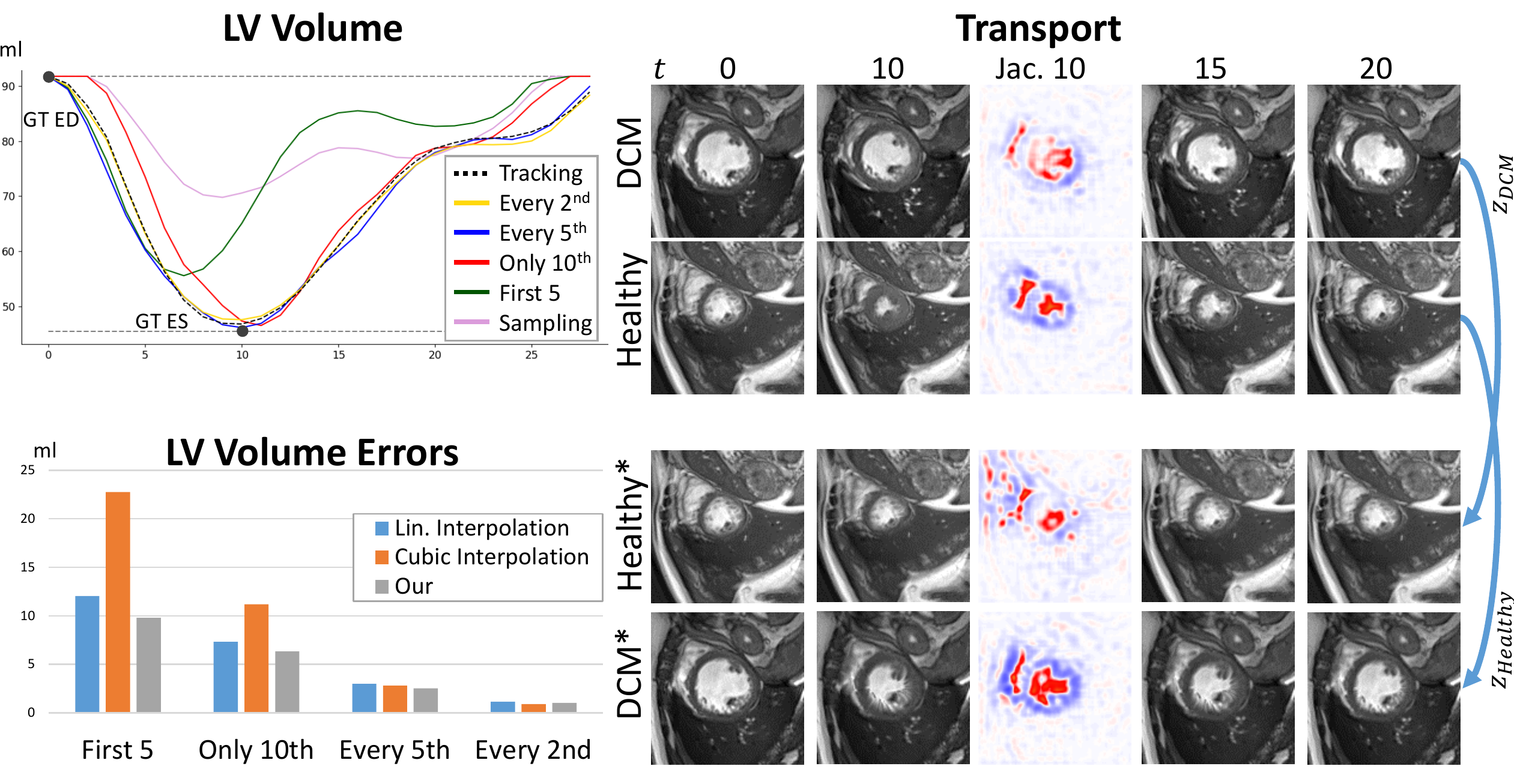}
%\vspace{-7pt}
\caption{\small{Left Top: LV volume curves from simulated and reconstructed motion by providing only a subset of frames and predicting the complete motion sequence from them. We provided only the first frame $I_0$ (sampling), every second, every fifth, the first five frames or only the tenths frame of the same sequence. Left Bottom: Mean volume errors with respect to the tracking volumes of all 677 testing cases comparing our sampling procedure with linear interpolation between velocity fields of the given frames. Right: The motion matrix $z$ from a pathological (first row: dilated myopathy DCM) and a healthy subject (second row) are transported from one to the other (bottom rows).}}\label{samp_images}
%\vspace{-8pt}
\end{figure}

\subsection{Sampling, Sequence Reconstruction and Motion Transport}
%\vspace{-2pt}
We then evaluate motion sampling, reconstruction and transport. We extract simulated reconstructed motion patterns if we provide our model with different subsets of images from the original sequence. In the time steps without frames, the motion matrix $z$ is created by randomly sampling $\tilde{z}_t$ from the prior distribution as depicted in Fig.~\ref{sampling_fig} (take same $\tilde{z}_t$ for all slices of one volume). Here, we choose sampling over interpolation in $\tilde{z}$-space to remain an uncertain, a probabilistic, estimation of the interpolated deformations. The left side of Fig.~\ref{samp_images} shows LV volume curves and reconstruction errors if every second, every fifth, only the 10th, no frame (sampling) or the first 5 frames are provided besides the moving frame (ED). The LV volume errors are computed on all test sequences by taking the mean absolute differences between sampled and tracking volumes. We compare with linear and cubic interpolation of velocities (extracted from tracking) between given times. One can see that our model performs better in recovering the LV volume, compared to linear and cubic interpolation when fewer frames are provided. Given the first five frames or only one additional frame (10th frame), the model estimates the motion consistently with plausible cardiac motion patterns for the missing time steps. In the cases of providing every second and every fifth frame, our method performs equally good or marginally better. Note, the motion simulation  given only the ED frame (sampling) does not overlap with the original motion, which is not intended. Nevertheless, one can see cardiac specific motion patterns such as the plateau phase before the atrial systole. 

Motion can be transported by taking the motion matrix $z$ from one sequence and apply it on the ED frame $I_0$ of another sequence. The right side of Fig.~\ref{samp_images} shows the transport of a pathological motion to a healthy subject and vice versa. The resulting simulated motion shows similar heart contractions and motion characteristics as the originating motion while the transported deformations are adapted to the target image without requiring explicit inter-subject registration. 

%Videos of motion transport and sampling  are available in the supplementary material.

%\vspace{-3pt}
\section{Conclusions}
%\vspace{-3pt}
In this paper, we presented an unsupervised approach for learning a motion model from image sequences. Underlying motion factors are encoded in a low-dimensional probabilistic space, the motion matrix, in which each column represents the deformation between two frames of the sequence. 
Our model demonstrated accurate motion tracking and motion reconstruction from missing frames, which can be useful for shorter acquisition times and temporal super-resolution. We also showed motion transport and simulation by using only one frame. Limitations of the presented approach include the support for 3-D image sequences and the generalization to other use cases such as respiratory motion. 

For future work, we aim to explore these points and especially the spatial coherence between slices for 3-D applications and the influence of using different training datasets (pathological and non-pathological) on the learned motion matrix.
\\

%\noindent \textbf{Acknowledgements: } Data used in this article were obtained from the EU FP7-funded project MD-Paedigree.

\noindent \textbf{Disclaimer: } The concepts and information presented in this paper are based on research results that are not commercially available.
%This feature is based on research, and is not commercially available. Due to regulatory reasons its future availability cannot be guaranteed.

%\vspace{-3pt}
\bibliographystyle{splncs04}

\bibliography{paper}

\begin{thebibliography}{10}
\providecommand{\url}[1]{\texttt{#1}}
\providecommand{\urlprefix}{URL }
\providecommand{\doi}[1]{https://doi.org/#1}

\bibitem{abadi2016tensorflow}
Abadi, M., Agarwal, A., Barham, P., Brevdo, E., Chen, Z., Citro, C., Corrado,
  G.S., Davis, A., Dean, J., Devin, M., et~al.: Tensorflow: Large-scale machine
  learning on heterogeneous distributed systems. arXiv preprint
  arXiv:1603.04467  (2016)

\bibitem{avants2008symmetric}
Avants, B.B., Epstein, C.L., Grossman, M., Gee, J.C.: Symmetric diffeomorphic
  image registration with cross-correlation: evaluating automated labeling of
  elderly and neurodegenerative brain. Medical image analysis  \textbf{12}(1),
  26--41 (2008)

\bibitem{koltun2018}
Bai, S., Kolter, J.Z., Koltun, V.: An empirical evaluation of generic
  convolutional and recurrent networks for sequence modeling. arXiv preprint
  arXiv:1803.01271  (2018)

\bibitem{bernard2018deep}
Bernard, O., Lalande, A., Zotti, C., Cervenansky, F., Yang, X., Heng, P.A.,
  Cetin, I., Lekadir, K., Camara, O., Ballester, M.A.G., et~al.: Deep learning
  techniques for automatic {MRI} cardiac multi-structures segmentation and
  diagnosis: Is the problem solved? IEEE Transactions on Medical Imaging
  \textbf{37}(11),  2514--2525 (2018)

\bibitem{caballero2017real}
Caballero, J., Ledig, C., Aitken, A., Acosta, A., Totz, J., Wang, Z., Shi, W.:
  Real-time video super-resolution with spatio-temporal networks and motion
  compensation. In: Proceedings of the IEEE Conference on Computer Vision and
  Pattern Recognition. pp. 4778--4787 (2017)

\bibitem{dalca2018unsupervised}
Dalca, A.V., Balakrishnan, G., Guttag, J., Sabuncu, M.R.: Unsupervised learning
  for fast probabilistic diffeomorphic registration. In: International
  Conference on Medical Image Computing and Computer-Assisted Intervention. pp.
  729--738. Springer (2018)

\bibitem{de2012temporal}
De~Craene, M., Piella, G., Camara, O., Duchateau, N., Silva, E., Doltra, A.,
  D’hooge, J., Brugada, J., Sitges, M., Frangi, A.F.: Temporal diffeomorphic
  free-form deformation: Application to motion and strain estimation from 3d
  echocardiography. Medical image analysis  \textbf{16}(2),  427--450 (2012)

\bibitem{kingma2014adam}
Kingma, D.P., Ba, J.: Adam: A method for stochastic optimization. arXiv
  preprint arXiv:1412.6980  (2014)

\bibitem{kingma2014semi}
Kingma, D.P., Mohamed, S., Rezende, D.J., Welling, M.: Semi-supervised learning
  with deep generative models. In: Advances in Neural Information Processing
  Systems. pp. 3581--3589 (2014)

\bibitem{krebs2019}
Krebs, J., Delingette, H., Mailh{\'e}, B., Ayache, N., Mansi, T.: Learning a
  probabilistic model for diffeomorphic registration. IEEE Transactions on
  Medical Imaging  \textbf{38} (2019)

\bibitem{lorenzi2013lcc}
Lorenzi, M., Ayache, N., Frisoni, G.B., Pennec, X., et~al.: {LCC-D}emons: a
  robust and accurate symmetric diffeomorphic registration algorithm.
  NeuroImage  \textbf{81},  470--483 (2013)

\bibitem{metz2011nonrigid}
Metz, C., Klein, S., Schaap, M., van Walsum, T., Niessen, W.J.: Nonrigid
  registration of dynamic medical imaging data using nd+ t b-splines and a
  groupwise optimization approach. Medical image analysis  \textbf{15}(2),
  238--249 (2011)

\bibitem{qin2018joint}
Qin, C., Bai, W., Schlemper, J., Petersen, S.E., Piechnik, S.K., Neubauer, S.,
  Rueckert, D.: Joint learning of motion estimation and segmentation for
  cardiac mr image sequences. In: International Conference on Medical Image
  Computing and Computer-Assisted Intervention. pp. 472--480. Springer (2018)

\bibitem{rohe2018low}
Roh{\'e}, M.M., Sermesant, M., Pennec, X.: Low-dimensional representation of
  cardiac motion using barycentric subspaces: A new group-wise paradigm for
  estimation, analysis, and reconstruction. Medical image analysis
  \textbf{45},  1--12 (2018)

\bibitem{shi2013temporal}
Shi, W., Jantsch, M., Aljabar, P., Pizarro, L., Bai, W., Wang, H., O’regan,
  D., Zhuang, X., Rueckert, D.: Temporal sparse free-form deformations. Medical
  image analysis  \textbf{17}(7),  779--789 (2013)

\bibitem{yang2011prediction}
Yang, L., Georgescu, B., Zheng, Y., Wang, Y., Meer, P., Comaniciu, D.:
  Prediction based collaborative trackers (pct): A robust and accurate approach
  toward 3d medical object tracking. IEEE transactions on medical imaging
  \textbf{30}(11),  1921--1932 (2011)

\end{thebibliography}

\end{document}